\documentclass[runningheads]{llncs}
\usepackage[T1]{fontenc}
\usepackage{hyperref}
\hypersetup{colorlinks}
\usepackage{xcolor}
\usepackage{mathtools}
\usepackage{bbm}
\usepackage{bm}
\usepackage{physics}
\usepackage[normalem]{ulem}
\usepackage{multirow}

\usepackage{graphicx}
\usepackage{color}

\begin{document}
\newcommand{\methodname}{MSKdeX}
\title{\methodname: Musculoskeletal (MSK) decomposition from an X-ray image for fine-grained estimation of lean muscle mass and muscle volume}

\titlerunning{\methodname}

\author{
  Yi Gu\inst{1,2} \and
  Yoshito Otake\inst{1} \and
  Keisuke Uemura\inst{3} \and
  Masaki Takao\inst{4} \and\\
  Mazen Soufi\inst{1} \and
  Yuta Hiasa\inst{1} \and
  Hugues Talbot\inst{2} \and
  Seiji Okata\inst{3} \and\\
  Nobuhiko Sugano\inst{5} \and
  Yoshinobu Sato\inst{1}
}

\authorrunning{Y. Gu, Y. Otake, et al.}
\institute{
  Division of Information Science, Graduate School of Science and Technology,\\Nara Institute of Science and Technology, Japan\\
  \email{\{gu.yi.gu4,otake,yoshi\}@is.naist.jp} \and
  CentraleSupélec, Université Paris-Saclay, France \and
  Department of Orthopaedics, Osaka University Graduate School of Medicine, Japan \and 
  Department of Bone and Joint Surgery,\\Ehime University Graduate School of Medicine, Japan \and
  Department of Orthopaedic Medical Engineering,\\Osaka University Graduate School of Medicine, Japan
}

\maketitle              
\begin{abstract}
Musculoskeletal diseases such as sarcopenia and osteoporosis are major obstacles to health during aging. Although dual-energy X-ray absorptiometry (DXA) and computed tomography (CT) can be used to evaluate musculoskeletal conditions, frequent monitoring is difficult due to the cost and accessibility (as well as high radiation exposure in the case of CT). We propose a method (named \methodname) to estimate fine-grained muscle properties from a plain X-ray image, a low-cost, low-radiation, and highly accessible imaging modality, through musculoskeletal decomposition leveraging fine-grained segmentation in CT. We train a multi-channel quantitative image translation model to decompose an X-ray image into projections of CT of individual muscles to infer the lean muscle mass and muscle volume. We propose the object-wise intensity-sum loss, a simple yet surprisingly effective metric invariant to muscle deformation and projection direction, utilizing information in CT and X-ray images collected from the \textit{same} patient. While our method is basically an unpaired image-to-image translation, we also exploit the nature of the bone's rigidity, which provides the paired data through 2D-3D rigid registration, adding strong pixel-wise supervision in unpaired training. Through the evaluation using a 539-patient dataset, we showed that the proposed method significantly outperformed conventional methods. The average Pearson correlation coefficient between the predicted and CT-derived ground truth metrics was increased from 0.460 to 0.863. We believe our method opened up a new musculoskeletal diagnosis method and has the potential to be extended to broader applications in multi-channel quantitative image translation tasks.

\keywords{Muscles \and Radiography \and Generative adversarial networks (GAN). \and Sarcopenia \and Image-to-image translation}
\end{abstract}

\begin{figure}
\includegraphics[width=\textwidth]{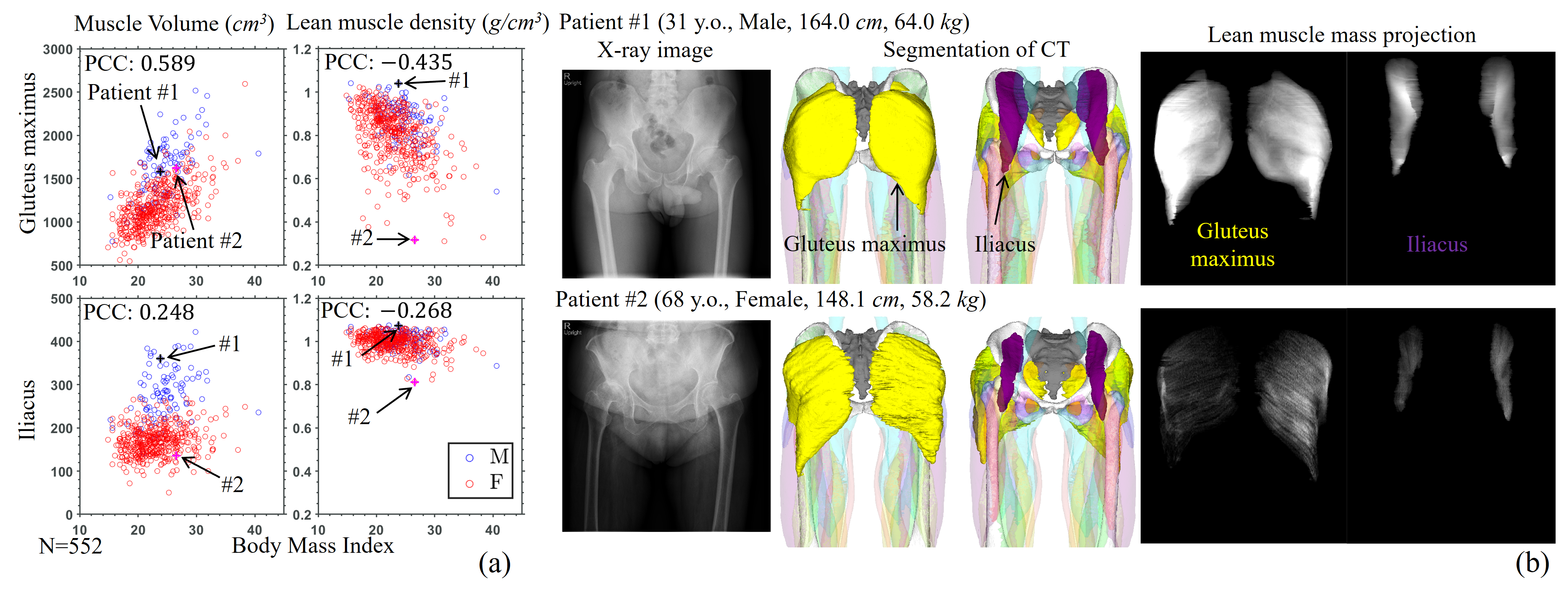}
\caption{Variations in the muscle volume and lean muscle density among the 552 patients in our dataset. 
(a) Relationships of muscle volume and lean muscle density with respect to body mass index (BMI). 
Moderate and weak correlations of BMI were observed with muscle volume and lean muscle density, respectively, in the gluteus maximus (Glu. max.), while little correlations were observed in the iliacus.
(b) Visualization of two representative cases. Patient \#1 (young, male) and Patient \#2 (old, female) had similar BMI and almost the same gluteus maximus volume, while the lean muscle mass was significantly different, likely due to the fatty degeneration in Patient \#2, which was clearly observable in the projections of the lean muscle mass volume.} 
\label{fig_intro}
\end{figure}
\section{Introduction}
Sarcopenia is a prevalent musculoskeletal disease characterized by the inevitable loss of skeletal muscle, causing increased risks of all-cause mortality and disability that result in heavy healthcare costs \cite{ref_kitamura_sarcopenia_2021,ref_chen_recent_2016,ref_marzetti_sarcopenia_2017,ref_petermann_global_2022,ref_edwards_osteoporosis_2015,ref_shu_diagnosis_2022}.
Measuring body composition, such as lean muscle mass (excluding fat contents), is essential for diagnosing musculoskeletal diseases, where dual-energy X-ray absorptiometry (DXA) \cite{ref_shepherd_body_2017,ref_nana_methodology_2014} and computed tomography (CT) \cite{ref_feliciano_evaluation_2020,ref_paris_automated_2020,ref_ogawa_validation_2020} are often used.
However, DXA and CT require special equipment that is much less accessible in a small clinic.
Furthermore, CT requires high radiation exposure, and DXA allows the measurement of only overall body composition, which lacks details in individual muscles such as the iliacus muscle, which overlays with the gluteus maximus muscle in DXA images.
Although several recent works used X-ray images for bone mineral density (BMD) estimation and osteoporosis diagnosis \cite{ref_hsieh_automated_2021,ref_wang_lumbar_2023,ref_ho_application_2021,ref_gu_bmd_2022}, only a few works estimated muscle metrics and sarcopenia diagnosis \cite{ref_ryu_chest_2023,ref_nakanishi_decomposition_2022}, and the deep learning technology used is old.
Recently, BMD-GAN \cite{ref_gu_bmd_2022} was proposed for estimating BMD through X-ray image decomposition using X-ray and CT images aligned by 2D-3D registration. 
However, they did not target muscles.
Nakanishi et al. \cite{ref_nakanishi_decomposition_2022} proposed an X-ray image decomposition for individual muscles. However, they calculated only the affected and unaffected muscle volumes ratio without considering the absolute volume and lean mass, which is more relevant to sarcopenia diagnosis.

In this study, we propose \textbf{\methodname}: Musculoskeletal (\textbf{MSK}) \textbf{de}composition from a plain \textbf{X}-ray image for the fine-grained estimation of lean muscle mass and volume of each individual muscle, which are useful metrics for evaluating muscle diseases including sarcopenia.
Fig. \ref{fig_intro} illustrates the meaning of our fine-grained muscle analysis and its challenges.
The contribution of this paper is three-fold: 1) proposal of the \textit{object-wise intensity-sum} (OWIS) loss, a simple yet effective metric invariant to muscle deformation and projection direction, for quantitative learning of the absolute volume and lean mass of the muscles, 2) proposal of partially aligned training utilizing the aligned (paired) dataset for the rigid object for the pixel-wise supervision in an unpaired image translation task, 3) extensive evaluation of the performance using a 539-patient dataset.

\begin{figure}
\includegraphics[width=\textwidth]{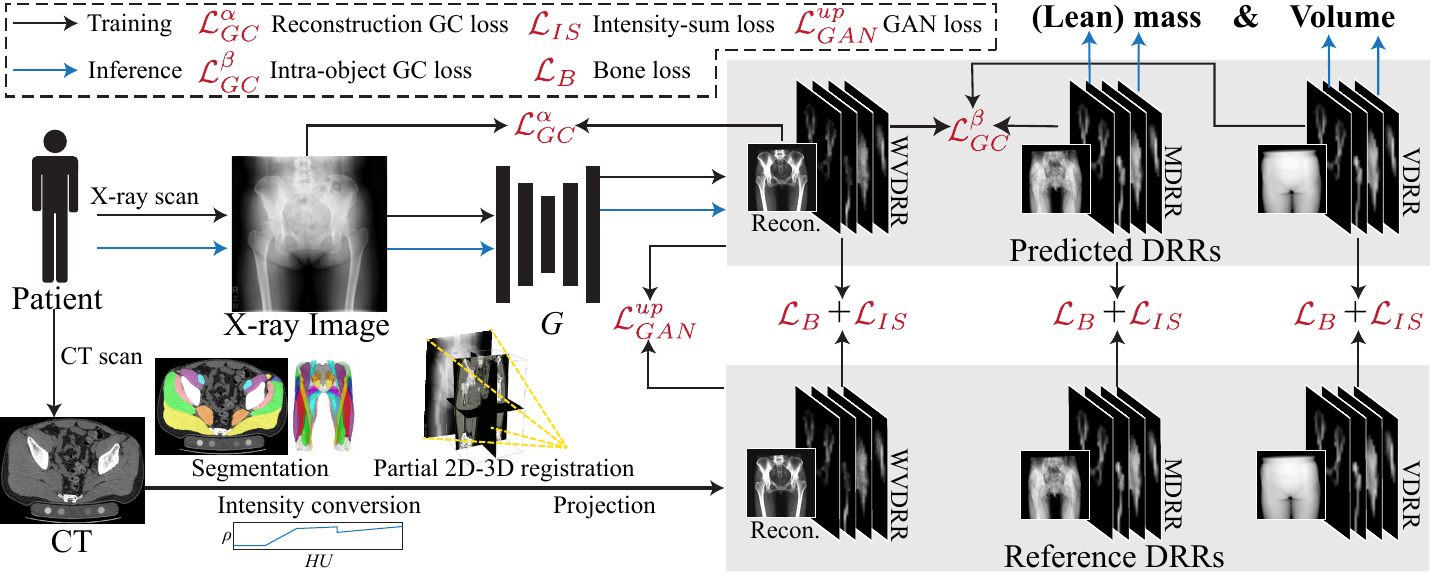}
\caption{Overview of the proposed \methodname.
Three types of object-wise DRRs (of segmented individual muscle/bone regions) were obtained from CT through segmentation \cite{ref_hiasa_automated_2019}, intensity conversion \cite{ref_aubrey_skeleta_muscle_2014,ref_wilfried_CT_numbers_2000}, 2D-3D registration for bones \cite{ref_otake_intraoperative_2012}, and projection, embedding information of volume and mass.
A decomposition model was trained using GAN loss and proposed GC loss chain, OWIS loss, and bone loss to decompose an X-ray image into DRRs whose intensity sum derives the metric of volume and mass.} 
\label{fig_method_overview}
\end{figure}
\section{Method}
\subsection{Dataset preparation}
\label{sec:dataset_preparation}
Fig. \ref{fig_method_overview} illustrates the overview of the proposed \methodname. We collected a dataset of 552 patients subject to the total hip arthroplasty surgery (455 females and 97 males, height 156.9 $\pm$ 8.3 cm, weight 57.5 $\pm$ 11.9 kg, BMI 23.294 $\pm$ 3.951 [mean $\pm$ std]). Ethical approval was obtained from the Institutional Review Boards of the institutions participating in this study (IRB approval numbers: 15056-3 for Osaka University and 2019-M-6 for Nara Institute of Science and Technology).
We acquired a pair of pre-operative X-ray and CT images from each patient, assuming consistency in bone shape, lean muscle mass, and muscle volume.
Automated segmentation of individual bones and muscles was obtained from CT \cite{ref_hiasa_automated_2019}.
Three different intensity conversions were applied to the segmented CT; 1) the original intensity, 2) intensity of 1.0 for voxels inside the structure and 0.0 for voxels outside to estimate muscle volume, 3) intensity corresponding to the lean muscle mass density based on a conversion function from the Hounsfield unit (HU) to the mass density \cite{ref_aubrey_skeleta_muscle_2014,ref_wilfried_CT_numbers_2000} to estimate lean muscle mass. Following \cite{ref_aubrey_skeleta_muscle_2014}, we assumed the voxels with less than -30 HU consisted of the fat, more than +30 HU consisted of the lean muscle, and the voxels in between -30 to +30 HU contained the fat and lean muscle with the ratio depending on linear interpolation of the HU value. The mass of the lean muscle was calculated by the conversion function proposed in \cite{ref_wilfried_CT_numbers_2000}.
Then, object-wise DRRs for the three conversions were generated for each segmented individual object (bone/muscle) region. 
(Note: When we refer to a DRR in this paper, it is object-wise.)
We call the three types of the DRRs weighted volume DRR (WVDRR), volume DRR (VDRR), and mass DRR (MDRR). The intensity sum of VDRR and MDRR amounts to each object's muscle volume ($cm^3$) and lean muscle mass ($g$), respectively. (Note: ``Muscle volume'' includes the fat in addition to lean muscle.) 
The summation of all the objects of WVDRRs becomes an image with a contrast similar to the real X-ray image used to calculate the reconstruction gradient correlation (GC) loss \cite{ref_nakanishi_decomposition_2022,ref_hiasa_cross_2018}.
A 2D-3D registration \cite{ref_otake_intraoperative_2012} of each bone between CT and X-ray image of the same patient was performed to obtain its DRR aligned with the X-ray image, which is used in the proposed partially aligned training. 

Since muscles deform depending on the joint angle, they are not aligned. 
Instead, we exploited the invariant property of muscles using the newly proposed intensity-sum loss.

\subsection{Model training}
We train a decomposition model $G$ to decompose an X-ray image into the $DRRs=\{VDRR, MDRR, WVDRR\}$ to infer the lean muscle mass and muscle volume, adopting CycleGAN \cite{ref_zhu_unpaired_2017}.
The model backbone is replaced with HRNet \cite{ref_wangg_deep_2021}.
The GAN loss $\mathcal{L}^{up}_{GAN}$ we use is formulated in supplemental materials.

\subsubsection{Structural consistency.}
We call the summation of a DRR over all the channels (objects) the virtual X-ray image defined as $I^{VX}=V(I^{DRR})=\sum_iI_i^{DRR}$, where $I_i^{DRR}$ is the $i$-th object image of a DRR.
We applies reconstruction GC loss \cite{ref_nakanishi_decomposition_2022} $\mathcal{L}^\alpha_{GC}$ defined as 
\begin{equation}
    \mathcal{L}^\alpha_{GC}(G)=\mathbbm{E}_{I^{X}}-GC(I^X,V(G(I^X)^{WVDRR}))
\end{equation}
to maintain the structure consistency between an X-ray image and decomposed DRR, where $G(I^X)^{WVDRR}$ is the decomposed WVDRR.
However, we do not apply reconstruction GC loss for VDRR and MDRR because of lacking attenuation coefficient information.
Instead, we propose inter-DRR/intra-object GC loss $\mathcal{L}^\beta_{GC}$ defined as
\begin{equation}
\begin{split}
    \mathcal{L}^\beta_{GC}(G)=\mathbbm{E}_{I^{X}}-\frac{1}{N}\sum_i\Bigl[GC(st(G(I^X)_i^{WVDRR}),G(I^X)_i^{VDRR})\\
    +GC(st(G(I^X)_i^{WVDRR}),G(I^X)_i^{MDRR})\Bigr]
\end{split}
\end{equation}
to chain the structural constraints from WVDRR to VDRR and MDRR, where the $G(I^X)_i^{WVDRR}$, $G(I^X)_i^{VDRR}$, and $G(I^X)_i^{MDRR}$ are $i$-th object image of the decomposed WVDRR, VDRR, and MDRR, respectively.
The $st(\cdot)$ operator stops the gradient from being back-propagated in which the decomposed VDRR and MDRR are expected to be structurally closer to WVDRR (not vice-versa) to stabilize training.
Thus, our structural consistency constant $\mathcal{L}^{up}_{GC}$ is defined as
$\mathcal{L}^{up}_{GC}=\lambda_{gca}\mathcal{L}^{\alpha}_{GC}(G)+\mathcal{L}^{\beta}_{GC}(G)$
where the $\lambda_{gca}$ balances the two GC losses. 

\subsubsection{Intensity sum consistency.}
Unlike general images, our DRRs embedded specific information so that the intensity sum represents physical metrics (mass and volume).
Furthermore, the conventional method did not utilize the paired information of an X-ray image and DRR (obtained from the same patient).
We took advantage of the paired information, proposing the \textit{object-wise intensity-sum} loss, a simple yet effective metric invariant to patient pose and projection direction, for quantitative learning.
The OWIS loss $\mathcal{L}_{IS}$ is defined as:
\begin{equation}
    \mathcal{L}_{IS}(DRR)=\mathbbm{E}_{(I^X,I^{DRR})}\frac{1}{NHW}\sum_i\Bigl|S(G(I^X)_i^{DRR})-S(I_i^{DRR})\Bigr|,
\end{equation}
where $I_i^{DRR}$ and $S(\cdot)$ are the $i$-th object image of DRR and the intensity summation operator (sum over the intensity of an image), respectively.
The $H$ and $W$ are the image height and weight, respectively, served as temperatures for numeric stabilizability.
The intensity consistency objective $\mathcal{L}^{all}_{IS}$ is defined as
$\mathcal{L}^{all}_{IS}=\mathcal{L}_{IS}(WVDRR)+\mathcal{L}_{IS}(VDRR)+\mathcal{L}_{IS}(MDRR)$.

\subsubsection{Partially aligned training.}
A previous study \cite{ref_gu_bmd_2022} suggested that supervision by the aligned (paired) data can improve the quantitative translation. 
Therefore, we incorporated 2D-3D registration \cite{ref_otake_intraoperative_2012} to align the pelvis and femur DRRs with the paired X-ray images for partially aligned training to improve overall performance, including muscle metrics estimation.
We applied $L1$ and GC loss to maintain quantitative and structural consistencies, respectively.
However, we preclude using GAN loss and \textit{feature matching} loss to avoid the training burden by additional discriminators.
The paired bone loss for a DRR is defined as
\begin{equation}
\begin{split}
    \mathcal{L}_{B}(DRR)=\mathbbm{E}_{(I^X,I^{DRR})}\frac{1}{N_b}\sum_{i\in K}\Bigl[\lambda_{l1}\norm{G(I^X)_i^{DRR}-I_i^{DRR}}_1\\
    -GC(G(I^X)_i^{DRR},I_i^{DRR})\Bigr],
\end{split}
\end{equation}
where the $K$ is a set of indexes containing aligned bone indexes.
The $N_b$ is the size of the set $K$.
The $\lambda_{l1}$ tries to balance structural faithfulness and quantitative accuracy.
The objective of partially aligned pixel-wise learning is defined as
$\mathcal{L}^{all}_B=\mathcal{L}_B(WVDRR)+\mathcal{L}_B(VDRR)+\mathcal{L}_B(MDRR)$.

\subsubsection{Full objective}
The full objective, aiming for realistic decomposition while maintaining structural faithfulness and quantitative accuracy, is defined as

\begin{equation}
\mathcal{L}=\min_{G,F}\max_{D^{X},D^{DRRs}}\biggl(\mathcal{L}^{up}_{GAN}+\mathcal{L}_{GC}+\lambda_{is}\mathcal{L}^{all}_{IS}+\mathcal{L}^{all}_B\biggr),
\end{equation}
where the $\lambda_{is}$ re-weights the penalty on the proposed OWIS loss.

\begin{figure}
\centering
\includegraphics[width=.992\textwidth]{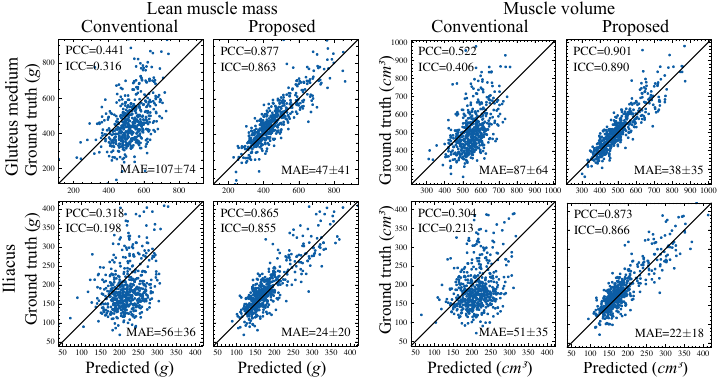}
\caption{Lean muscle mass (left) and muscle volume (right) estimation results by the conventional and proposed methods for the gluteus medius (top) and iliacus (bottom).} \label{fig_mass_volume_scatter}
\end{figure}

\begin{figure}
\includegraphics[width=\textwidth]{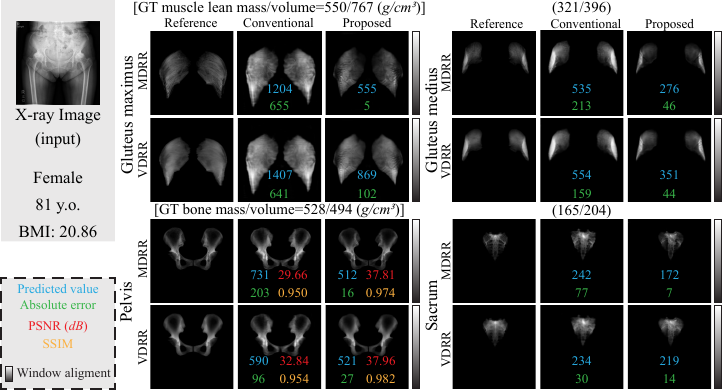}
\caption{Visualization of decomposition results of the MDRR and VDRR for gluteus maximus, gluteus medius, pelvis, and sacrum on a representative case.} \label{fig_visual_pred}
\end{figure}
\section{Experiments and Results}
\label{sec:results}
The automatic segmentation results of 552 CTs were visually verified, and 13 cases with severe segmentation failures were omitted from our analysis, resulting in 539 CTs. Four-fold cross-validation was performed, i.e., 404 or 405 training data and 134 or 135 test data per fold.
The baseline of our experiment was the vanilla CycleGAN with the reconstruction GC loss proposed in \cite{ref_nakanishi_decomposition_2022}. We evaluated the predicted lean muscle mass and volume using the ground truth derived from 3D CT images with three metrics, Pearson correlation coefficient (PCC), intra-class correlation coefficient (ICC), and mean absolute error (MAE). 
Additionally, we evaluated the image quality of predicted DRRs of the bones by comparing them with the aligned DRRs using peak-signal-noise-ratio (PSNR) and structural similarity index measure (SSIM).
Implementation details are described in supplemental materials.

Fig. \ref{fig_mass_volume_scatter} shows the prediction results on the gluteus medius and iliacus muscles.
The conventional method (without using the intensity constraints) resulted in low PCCs of 0.441 and 0.522 for the lean muscle mass and muscle volume estimations, respectively, for the gluteus medius, and 0.318 and 0.304, respectively, for the iliacus.
Significant improvements by the proposed method were observed, achieving high PCCs of 0.877 and 0.901 of the lean muscle mass and muscle volume estimations, respectively, for the gluteus medius, and 0.865 and 0.873, respectively, for the iliacus.
Fig. \ref{fig_visual_pred} visualized the decomposed VDRR and MDRR of four objects of a representative case.
Our method (\methodname) reduced the hallucinating features in the decomposed DRRs by the proposed losses.
The overall intensity of the conventional method was clearly different from the reference, while the proposed method decomposed the X-ray image considering the structural faithfulness and quantitative accuracy, outperforming the conventional method significantly. 
Table \ref{tab:main_eval} shows evaluation for other objects. Statistical test (one-way ANOVA) was performed on the conventional and proposed methods using prediction absolute error, where the differences are significant ($p<0.001$) for all the objects. More detailed results and a visualization video can be found in supplemental materials.

\subsubsection{Ablation study.}
\label{sec:ablation}
We performed ablation studies to investigate the impact of proposed OWIS loss and the use of aligned bones using 404 training and 135 test data. 
The re-weighting parameter $\lambda_{is}$ of 0, 10, and 1000 with and without the partially aligned training $\mathcal{L}_{B}$ was tested. The $\lambda_{is}$ of 0 without partially aligned training [$\lambda_{is}=0$ (False)] is considered our baseline.
The results of the ablation study were summarized in Table \ref{tab:abla}, where the bold font indicated the best setting in a column.
We observed significant improvements from the baseline by both proposed features, OWIS loss and partially aligned training $\mathcal{L}_{B}$.

The average PCC for the muscles was improved from 0.457 to 0.826 by adding the OWIS loss ($\lambda_{is}=100$) and to 0.796 by adding the bone loss, while their combination achieved the best average PCC of 0.855, demonstrating the superior ability of quantitative learning of the proposed \methodname. 
The results also suggested that the weight balance for loss terms needs to be made to achieve the best performance.
More detailed results are shown in supplemental materials.

\begin{table}[]
\caption{Performance comparison between the conventional and proposed methods in a cross-validation study using 539 data. 
}
\centering
\begin{tabular}{|cclclclcccccc|}
\hline
\multicolumn{13}{|c|}{Lean muscle mass and Bone mass estimation accuracy (PCC)}                                                                                                                                                                                                                                 \\ \hline
\multicolumn{1}{|c|}{Method} & \multicolumn{2}{c|}{Glu. max.}     & \multicolumn{2}{c|}{Glu. med.}     & \multicolumn{2}{c|}{Glu. min.}     & \multicolumn{1}{c|}{Ilia.}         & \multicolumn{2}{c|}{Obt. ext.}     & \multicolumn{1}{c|}{Pec.}          & \multicolumn{1}{c|}{Pelv.}         & Sac.          \\ \hline
\multicolumn{1}{|c|}{Conv.}  & \multicolumn{2}{c|}{.417}          & \multicolumn{2}{c|}{.441}          & \multicolumn{2}{c|}{.415}          & \multicolumn{1}{c|}{.318}          & \multicolumn{2}{c|}{.416}          & \multicolumn{1}{c|}{.457}          & \multicolumn{1}{c|}{.612}          & .478          \\
\multicolumn{1}{|c|}{Prop.}  & \multicolumn{2}{c|}{\textbf{.831}} & \multicolumn{2}{c|}{\textbf{.877}} & \multicolumn{2}{c|}{\textbf{.864}} & \multicolumn{1}{c|}{\textbf{.865}} & \multicolumn{2}{c|}{\textbf{.825}} & \multicolumn{1}{c|}{\textbf{.832}} & \multicolumn{1}{c|}{\textbf{.950}} & \textbf{.878} \\ \hline
\multicolumn{13}{|c|}{Muscle volume and bone volume estimation accuracy (PCC)}                                                                                                                                                                                                                                  \\ \hline
\multicolumn{1}{|c|}{Method} & \multicolumn{2}{c|}{Glu. max.}     & \multicolumn{2}{c|}{Glu. med.}     & \multicolumn{2}{c|}{Glu. min.}     & \multicolumn{1}{c|}{Ilia.}         & \multicolumn{2}{c|}{Obt. ext.}     & \multicolumn{1}{c|}{Pec.}          & \multicolumn{1}{c|}{Pelv.}         & Sac.          \\ \hline
\multicolumn{1}{|c|}{Conv.}  & \multicolumn{2}{c|}{.490}          & \multicolumn{2}{c|}{.522}          & \multicolumn{2}{c|}{.463}          & \multicolumn{1}{c|}{.304}          & \multicolumn{2}{c|}{.402}          & \multicolumn{1}{c|}{.460}          & \multicolumn{1}{c|}{.586}          & .538          \\
\multicolumn{1}{|c|}{Prop.}  & \multicolumn{2}{c|}{\textbf{.882}} & \multicolumn{2}{c|}{\textbf{.901}} & \multicolumn{2}{c|}{\textbf{.890}} & \multicolumn{1}{c|}{\textbf{.873}} & \multicolumn{2}{c|}{\textbf{.864}} & \multicolumn{1}{c|}{\textbf{.849}} & \multicolumn{1}{c|}{\textbf{.956}} & \textbf{.873} \\ \hline
\multicolumn{13}{|c|}{Bone decomposition accuracy {[}mean PSNR(mean SSIM){]}}                                                                                                                                                                                                                                   \\ \hline
\multicolumn{1}{|c|}{Method} & \multicolumn{4}{c|}{MDRR Pelv., Fem.}                                   & \multicolumn{4}{c|}{VDRR  Pelv., Fem.}                                                     & \multicolumn{4}{c|}{MVDRR  Pelv., Fem.}                                                                   \\ \hline
\multicolumn{1}{|c|}{Conv.}  & \multicolumn{4}{c|}{31.5(.940), 31.7(.962)}                             & \multicolumn{4}{c|}{31.7(.940), 32.4(.963)}                                                & \multicolumn{4}{c|}{34.5(.956), 31.6(.971)}                                                               \\
\multicolumn{1}{|c|}{Prop.}  & \multicolumn{4}{c|}{\textbf{37.1(.978), 37.8(.987)}}                    & \multicolumn{4}{c|}{\textbf{37.7(.982), 39.5(.991)}}                                       & \multicolumn{4}{c|}{\textbf{39.0(.980), 37.1(.987)}}                                                      \\ \hline
\end{tabular}
\label{tab:main_eval}
\end{table}
\section{Summary}
We proposed \methodname, a method for fine-grained estimation of the lean muscle mass and volume from a plain X-ray image (2D) through the musculoskeletal decomposition, which, in fact, recovers CT (3D) information. 
Our method decomposes an X-ray image into DRRs of objects to infer the lean muscle mass and volume considering the structural faithfulness (by the gradient correlation loss chain) and quantitative accuracy (by the object-wise intensity-sum loss and aligned bones training), outperforming the conventional method by a large margin as shown in Sec. \ref{sec:results}.
The results suggested a high potential of \methodname{ }for opportunistic screening of musculoskeletal diseases in routine clinical practice, providing a new approach to accurately monitoring musculoskeletal health.
The aligned bone DRRs positively affected the quantification of the density and volume of the muscles as shown in the ablation study in Sec. \ref{sec:ablation}, implying the deep connection between muscles and bones. 
The prediction of muscles overlapped with the pelvis in the X-ray image can leverage the strong pixel-wise supervision by the aligned pelvis's DRR, which can be considered as a type of \textit{calibration}.
Our future works are the validation with a large-scale dataset and extension to the decomposition into a larger number of objects.
\begin{table}
\caption{Summary of the ablation study for 135 test data. 
}
\centering
\begin{tabular}{|ll|cccccccc|}
\hline
\multirow{2}{*}{$\lambda_{is}$} & \multirow{2}{*}{With $\mathcal{L}_{B}$} & \multicolumn{8}{c|}{Lean muscle mass and bone mass estimation accuracy (PCC)}                                                                                                                                                                                                    \\ \cline{3-10} 
                        &                              & \multicolumn{1}{c|}{Glu. max.}     & \multicolumn{1}{c|}{Glu. med.}     & \multicolumn{1}{c|}{Glu. min.}     & \multicolumn{1}{c|}{Ilia.}         & \multicolumn{1}{c|}{Obt. ext.}     & \multicolumn{1}{c|}{Pec.}          & \multicolumn{1}{c|}{Pelv.}         & Sac.          \\ \hline
0                       & (False)                      & \multicolumn{1}{c|}{.415}          & \multicolumn{1}{c|}{.419}          & \multicolumn{1}{c|}{.469}          & \multicolumn{1}{c|}{.368}          & \multicolumn{1}{c|}{.473}          & \multicolumn{1}{c|}{.600}          & \multicolumn{1}{c|}{.542}          & .265          \\
0                       & (True)                       & \multicolumn{1}{c|}{.734}          & \multicolumn{1}{c|}{.799}          & \multicolumn{1}{c|}{.788}          & \multicolumn{1}{c|}{.855}          & \multicolumn{1}{c|}{.784}          & \multicolumn{1}{c|}{.813}          & \multicolumn{1}{c|}{\textbf{.954}} & .798          \\
100                     & (False)                      & \multicolumn{1}{c|}{.799}          & \multicolumn{1}{c|}{.815}          & \multicolumn{1}{c|}{.829}          & \multicolumn{1}{c|}{.854}          & \multicolumn{1}{c|}{.815}          & \multicolumn{1}{c|}{.842}          & \multicolumn{1}{c|}{.925}          & .774          \\
100                     & (True)                       & \multicolumn{1}{c|}{\textbf{.854}} & \multicolumn{1}{c|}{\textbf{.857}} & \multicolumn{1}{c|}{\textbf{.837}} & \multicolumn{1}{c|}{\textbf{.883}} & \multicolumn{1}{c|}{\textbf{.854}} & \multicolumn{1}{c|}{\textbf{.846}} & \multicolumn{1}{c|}{.947}          & \textbf{.898} \\
1000                    & (False)                      & \multicolumn{1}{c|}{.798}          & \multicolumn{1}{c|}{.765}          & \multicolumn{1}{c|}{.770}          & \multicolumn{1}{c|}{.839}          & \multicolumn{1}{c|}{.704}          & \multicolumn{1}{c|}{.840}          & \multicolumn{1}{c|}{.870}          & .767          \\
1000                    & (True)                       & \multicolumn{1}{c|}{.795}          & \multicolumn{1}{c|}{.776}          & \multicolumn{1}{c|}{.767}          & \multicolumn{1}{c|}{.828}          & \multicolumn{1}{c|}{.748}          & \multicolumn{1}{c|}{.820}          & \multicolumn{1}{c|}{.929}          & .745          \\ \hline
\end{tabular}
\label{tab:abla}
\end{table}
\subsubsection{Acknowledgement.}
The research in this paper was funded by\\
MEXT/JSPS KAKENHI (19H01176, 20H04550, 21K16655).
\subsubsection{Code availability.}
The code is available from the authors (\{gu.yi.gu4,otake, yoshi\}@is.naist.jp) upon reasonable request for research activity.
%
%
%
%

\end{document}


\newcommand{\methodname}{MSKdeX}
\newcommand{\RomanNumeralCaps}[1]
    {\MakeUppercase{\romannumeral #1}}
\title{Supplemental materials for \\\methodname: Musculoskeletal (MSK) decomposition from an X-ray image for fine-grained estimation of lean muscle mass and muscle volume}
\titlerunning{\methodname}
%
%
\authorrunning{************}
%

%

%
%
%

\section*{Supplemental materials for\\\methodname: Musculoskeletal (MSK) decomposition from an X-ray image for fine-grained estimation of lean muscle mass and muscle volume}

Note: the reference number corresponds to the one in the main paper.

\begin{table}
\caption{Results summary of ICC by the conventional and proposed methods for the muscles of gluteus maximus (glu. max.), gluteus medius (glu. med.), gluteus minimus (glu. min.), iliacus (ilia.), obturator externus (obt. ext.), and pectineus (pec.) and the bones of pelvis (pelv.) and sacrum (sac.). 
Similar to PCC shown in Table 1 in the main paper, the proposed method, which uses the object-wise intensity-sum (OWIS) loss and bone loss through partially aligned training, showed superior performance by a large margin compared to the conventional method, which uses only structural consistency supervision.
}
\centering
\begin{tabular}{|c|cccccccc|}
\hline
       & \multicolumn{8}{c|}{Lean muscle mass and bone mass estimation accuracy}                                                                                                                                                                                                                  \\ \hline
Method & \multicolumn{1}{c|}{Glu. max.}      & \multicolumn{1}{c|}{Glu. med.}      & \multicolumn{1}{c|}{Glu. min.}      & \multicolumn{1}{c|}{Ilia.}          & \multicolumn{1}{c|}{Obt. ext.}      & \multicolumn{1}{c|}{Pec.}           & \multicolumn{1}{c|}{Pelv.}          & Sac.           \\ \hline
Conv.  & \multicolumn{1}{c|}{0.310}          & \multicolumn{1}{c|}{0.316}          & \multicolumn{1}{c|}{0.372}          & \multicolumn{1}{c|}{0.198}          & \multicolumn{1}{c|}{0.359}          & \multicolumn{1}{c|}{0.334}          & \multicolumn{1}{c|}{0.517}          & 0.309          \\
Prop.  & \multicolumn{1}{c|}{\textbf{0.758}} & \multicolumn{1}{c|}{\textbf{0.863}} & \multicolumn{1}{c|}{\textbf{0.853}} & \multicolumn{1}{c|}{\textbf{0.855}} & \multicolumn{1}{c|}{\textbf{0.820}} & \multicolumn{1}{c|}{\textbf{0.825}} & \multicolumn{1}{c|}{\textbf{0.936}} & \textbf{0.848} \\ \hline
       & \multicolumn{8}{c|}{Muscle volume and bone volume estimation accuracy}                                                                                                                                                                                                                   \\ \hline
Conv.  & \multicolumn{1}{c|}{0.375}          & \multicolumn{1}{c|}{0.406}          & \multicolumn{1}{c|}{0.424}          & \multicolumn{1}{c|}{0.213}          & \multicolumn{1}{c|}{0.333}          & \multicolumn{1}{c|}{0.344}          & \multicolumn{1}{c|}{0.494}          & 0.423          \\

rop.  & \multicolumn{1}{c|}{\textbf{0.820}} & \multicolumn{1}{c|}{\textbf{0.890}} & \multicolumn{1}{c|}{\textbf{0.882}} & \multicolumn{1}{c|}{\textbf{0.866}} & \multicolumn{1}{c|}{\textbf{0.858}} & \multicolumn{1}{c|}{\textbf{0.840}} & \multicolumn{1}{c|}{\textbf{0.950}} & \textbf{0.858} \\ \hline
\end{tabular}
\end{table}
\begin{figure}
\includegraphics[width=\textwidth]{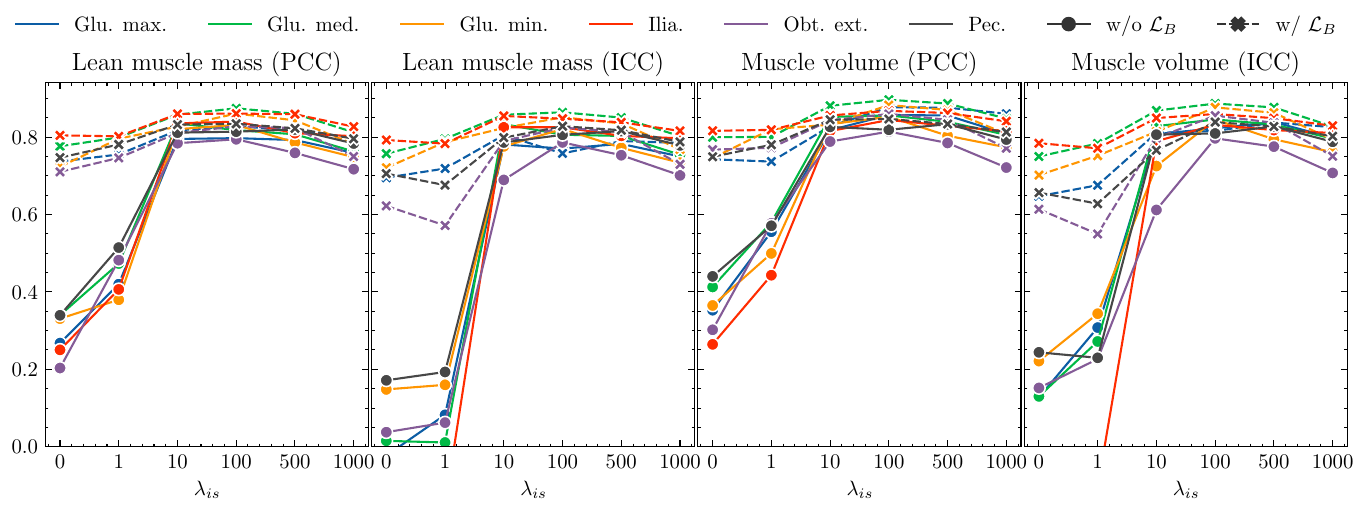}
\caption{Results of the tuning of the proposed OWIS loss ($\lambda_{is}$) and ablation of the bone loss ($\mathcal{L}_{B}$) for evaluating their impact using four-fold cross-validation (539 data). The bone loss was effective for any $\lambda_{is}$ and $\lambda_{is}=100$ resulting in the highest correlation for both lean muscle mass and muscle volume estimations.}
\end{figure}

\begin{figure}
\centering
\includegraphics[width=\textwidth]{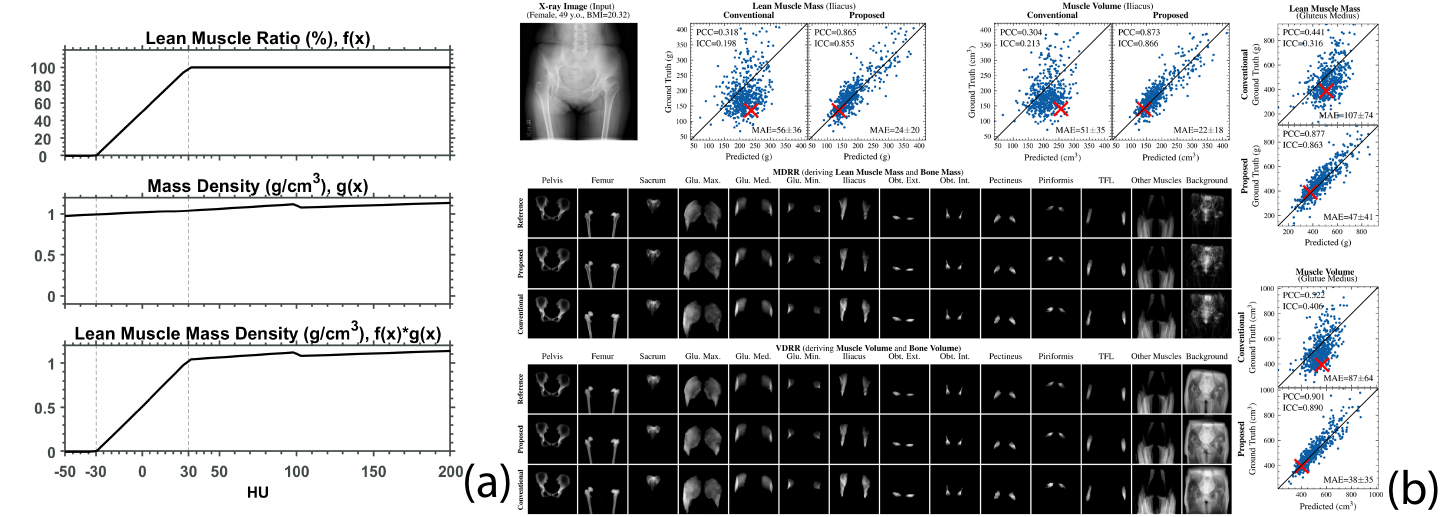}
\caption{CT intensity conversion method and the visualization of detailed results. 
(a) Conversion function from HU to lean muscle mass density (bottom), which is defined as the multiplication of two conversion functions about muscle/fat ratio (top) [19] and mass density (middle) [20].
(b) A screenshot of the supplemental video showing decomposed DRRs for all cases.}
\end{figure}
\subsubsection{Objective $\lambda^{up}_{GAN}$.}
Following CycleGAN [23], the GAN loss is defined as 
\begin{equation}
    \mathcal{L}_{GAN}(G,D,X,Y)=\mathbbm{E}_X[\log D(Y)] + \mathbbm{E}_Y[\log (1-D(G(X)))],
\end{equation}
where $G$, $D$, $X$, and $Y$ are the generator, discriminator, source data, and target data, respectively.
The \textit{cycle consistency} loss $\mathcal{L}_{cyc}$ [23] is defined as
\begin{equation}
    \mathcal{L}_{cyc}(G,F)=\mathbbm{E}_{I^{X}}[\norm{F(G(I^{X}))-I^{X}}_1]+\mathbbm{E}_{I^{DRRs}}[\norm{G(F(I^{DRRs}))-I^{DRRs}}_1],
\end{equation}
where $F$ is a twine generator to $G$, trying to construct an X-ray image from given DRRs.
The objective of the unpaired GAN is defined as
$\mathcal{L}^{up}_{GAN}=\lambda_{cyc}\mathcal{L}_{cyc}(G,F)+\mathcal{L}_{GAN}(G,D^{DRRs},I^X,I^{DRRs})+\mathcal{L}_{GAN}(F,D^{X},I^X,I^{DRRs})$,
where the $D^{DRRs}$ and $D^{X}$ are discriminators for DRRs and X-ray images, respectively.
The $\lambda_{cyc}$ determines the importance of the \textit{cycle consistency} loss.
\begin{table}
\caption{Implementation details}\label{tab1}
\centering
\begin{tabular}{|cccccc|}
\hline
\multicolumn{6}{|c|}{Part \RomanNumeralCaps{1}}                                                                                                                                                                                                                                                                                                                                                                                                                                                                                                    \\ \hline
\multicolumn{1}{|c|}{Study}      & \multicolumn{1}{c|}{\begin{tabular}[c]{@{}c@{}}Total \\ epochs\end{tabular}} & \multicolumn{1}{c|}{Optimizer}                                                                             & \multicolumn{1}{c|}{\begin{tabular}[c]{@{}c@{}}Learning rate \\ (LR) policy\end{tabular}}           & \multicolumn{1}{c|}{\begin{tabular}[c]{@{}c@{}}Inital LR,\\ min LR\end{tabular}}           & \begin{tabular}[c]{@{}c@{}}Balance\\ parameters\end{tabular}                                  \\ \hline
\multicolumn{1}{|c|}{Cross-val.} & \multicolumn{1}{c|}{600}                                                     & \multicolumn{1}{c|}{\multirow{2}{*}{\begin{tabular}[c]{@{}c@{}}AdamW (weight \\ decay=$1e^{-4}$)\end{tabular}}} & \multicolumn{1}{c|}{\multirow{2}{*}{\begin{tabular}[c]{@{}c@{}}SGDR [*]\\($T_0$=200, $T_i$=1)\end{tabular}}} & \multicolumn{1}{c|}{\multirow{2}{*}{\begin{tabular}[c]{@{}c@{}}$1e^{-4}$,\\ $1e^{-6}$\end{tabular}}} & \multirow{2}{*}{\begin{tabular}[c]{@{}c@{}}$\lambda_{cyc}$=10, $\lambda_{gca}$=0.5\\ $\lambda_{l1}$=100, $\lambda_{is}$=100\end{tabular}} \\ \cline{1-2}
\multicolumn{1}{|c|}{Ablation}   & \multicolumn{1}{c|}{400}                                                     & \multicolumn{1}{c|}{}                                                                                      & \multicolumn{1}{c|}{}                                                                               & \multicolumn{1}{c|}{}                                                                      &                                                                                               \\ \hline
\multicolumn{6}{|c|}{Part \RomanNumeralCaps{2}}                                                                                                                                                                                                                                                                                                                                                                                                                                                                                                    \\ \hline
\multicolumn{2}{|c|}{Method}                                                                                    & \multicolumn{1}{c|}{Time cost}                                                                             & \multicolumn{1}{c|}{Image size}                                                                     & \multicolumn{1}{c|}{Batch size}                                                            & GPU                                                                                           \\ \hline
\multicolumn{2}{|c|}{Conv.}                                                                                     & \multicolumn{1}{c|}{3 min/epoch}                                                                           & \multicolumn{1}{c|}{\multirow{2}{*}{512x512}}                                                       & \multicolumn{1}{c|}{\multirow{2}{*}{2}}                                                    & \multirow{2}{*}{RTX A6000}                                                                    \\ \cline{1-3}
\multicolumn{2}{|c|}{Prop.}                                                                                     & \multicolumn{1}{c|}{4 min/epoch}                                                                           & \multicolumn{1}{c|}{}                                                                               & \multicolumn{1}{c|}{}                                                                      &                                                                                               \\ \hline
\multicolumn{6}{l}{\multirow{2}{12cm}{\small [*] Loshchilov I., Hutter, F.: SGDR: Stochastic gradient descent with warm restarts. ICLR (2017)}}
\end{tabular}

\end{table}